%% file: main.tex
\documentclass[sigconf]{acmart}
\usepackage{hyperref}
\usepackage{makecell}
\usepackage{multirow}
\usepackage{subcaption}
\usepackage{graphicx}
\usepackage{tabularx}
\usepackage{url}
\usepackage{enumitem}
\usepackage{array}
\usepackage{booktabs}
\usepackage{amsmath}
\usepackage{pifont}
\usepackage{mdframed}
\usepackage{pifont} % For \xmark

\newcolumntype{Z}{>{\raggedright\arraybackslash}X} % Left-align text and treat as X
\newcolumntype{S}{>{\hsize=.4\hsize}Z} 
\newcolumntype{T}{>{\hsize=.8\hsize}Z} 

\newcommand\itemtitle[1]{\textcolor{orange}{#1}}
\newcommand\intent[1]{\textcolor{teal}{#1}}
\newcommand{\autoeval}[1]{\colorbox{yellow}{#1}}
\newcommand{\humaneval}[1]{\colorbox{pink}{#1}}

\newcommand{\success}{\ding{51}}
\newcommand{\failure}{\ding{55}}

\newcommand{\myvspace}{\vspace{0.1em}}

\newcommand{\boldheading}[1]{%
    \vspace{0.5em} 
    \noindent\textbf{#1}\hspace{0.1em} 
}

\AtBeginDocument{%
  \providecommand\BibTeX{{%
    \normalfont B\kern-0.5em{\scshape i\kern-0.25em b}\kern-0.8em\TeX}}}

\copyrightyear{2024}
\acmYear{2024}
\setcopyright{rightsretained}
\acmConference[CIKM '24]{Proceedings of the 33rd ACM International Conference on Information and Knowledge Management}{October 21--25, 2024}{Boise, ID, USA}
\acmBooktitle{Proceedings of the 33rd ACM International Conference on Information and Knowledge Management (CIKM '24), October 21--25, 2024, Boise, ID, USA}
\acmDOI{10.1145/3627673.3679928}
\acmISBN{979-8-4007-0436-9/24/10}

\makeatletter
\gdef\@copyrightpermission{
 \begin{minipage}{0.3\columnwidth}
 \href{https://creativecommons.org/licenses/by/4.0/}{\includegraphics[width=0.90\textwidth]{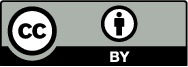}}
 \end{minipage}\hfill
 \begin{minipage}{0.7\columnwidth}
 \href{https://creativecommons.org/licenses/by/4.0/}{This work is licensed under a Creative Commons
Attribution International 4.0 License.}
 \end{minipage}
 \vspace{5pt}
}
\makeatother

\begin{document}

% \title{Predicting Live Chat Intent from Browsing History}
% \title{Predicting Intent from Live Chat Browsing History}
\title{Forecasting Live Chat Intent from Browsing History}

\author{Se-eun Yoon}
\affiliation{%
  \institution{University of California, San Diego}
  \city{La Jolla}
  \state{CA}
  \country{USA}}
\email{seeuny@ucsd.edu}

\author{Ahmad Bin Rabiah}
\affiliation{%
  \institution{University of California, San Diego}
  \city{La Jolla}
  \state{CA}
  \country{USA}}
\email{abinrabiah@ucsd.edu}

\author{Zaid Alibadi}
\affiliation{%
  \institution{Lowe’s Companies, Inc.}
  \city{Mooresville}
  \state{NC}
  \country{USA}}
\email{zalibadi@email.sc.edu}

\author{Surya Kallumadi}
\affiliation{%
  \institution{Lowe’s Companies, Inc.}
  \city{Mooresville}
  \state{NC}
  \country{USA}}
\email{surya@ksu.edu}

\author{Julian McAuley}
\affiliation{%
  \institution{University of California, San Diego}
  \city{La Jolla}
  \state{CA}
  \country{USA}}
\email{jmcauley@ucsd.edu}

\renewcommand{\shortauthors}{Se-eun Yoon, Ahmad Bin Rabiah, Zaid Alibadi, Surya Kallumadi, \& Julian McAuley}

\begin{abstract}
Customers reach out to online live chat agents with various intents, such as asking about product details or requesting a return.
In this paper, we propose the problem of predicting user intent from browsing history and address it through a two-stage approach.
The first stage classifies a user's browsing history into high-level intent categories.
Here, we represent each browsing history as a text sequence of page attributes and use the ground-truth class labels to fine-tune pretrained Transformers.
The second stage provides a large language model (LLM) with the browsing history and predicted intent class to generate fine-grained intents.
For automatic evaluation, we use a separate LLM to judge the similarity between generated and ground-truth intents, which closely aligns with human judgments.
Our two-stage approach yields significant performance gains compared to generating intents without the classification stage.
\end{abstract}



% \ccsdesc[500]{Information systems~Web searching and information discovery}

% \ccsdesc[500]{Information systems~Users and interactive retrieval}




\maketitle

\input{sections/introduction}
\input{sections/dataset}
\input{sections/methods}
\input{sections/results}
\input{sections/related}
\input{sections/conclusion}

\bibliographystyle{ACM-Reference-Format}
\bibliography{main}

% \appendix
% \input{appendix.tex}

\end{document}

%% file: sections/introduction.tex
\begin{figure}[t]
    \centering
    \vspace{1em}
    \includegraphics[width=0.85\linewidth]{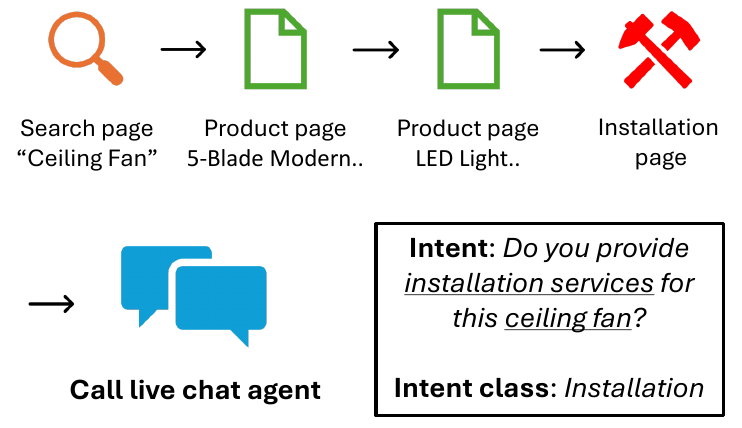}
    \caption{Example of a user browsing pages in an online store and then calling for a live chat agent. Out goal is to predict the user's intent, which is the reason for requesting assistance.}
    \label{fig:figure1}
    % \vspace{-1em}
\end{figure}

\section{Introduction}

Online business platforms often deploy live chat agents.
These agents address a wide variety of user intents, ranging from item availability to returns.
Even within the same broad topic, user intents can be highly specific:
one may ask how to install a certain item; another may ask whether installation services are included with the purchase.
Identifying each intent often requires several interactions in the chat interface, where users first navigate through a series of options to categorize their topic, and further clarify their needs once connected with a live agent.
% Identifying each intent often requires initial interactions, such as navigating through various options and providing explanations, adding complexity to both users and businesses.
% Many users are confused or frustrated by the time they reach out to an agent; some of them may bypass the service due to unawareness or due to the perceived hassle.

In this paper, we investigate whether it is possible to predict user intent \textit{before} users reach out to a live chat agent.
Solving this problem offers practical benefits.
The predicted intents can be used to automatically route users to specialized agents or prioritize wait times.
% The predicted intents may be used for automatic routing of users to specialized agents;
% they could also help to prioritize customers in wait times~\cite{goes2012live}.
% or, a targeted pop-up message may enhance user convenience by suggesting the possible topic that the user is struggling with.
With such applications in mind, we propose to predict user intent, specifically through users' browsing histories.
Browsing histories contain signals of intent~\cite{li2023text, guo2019buying}.
Each page visit has attributes such as page content, which hint at a user's specific intent.

Our problem is depicted in Figure~\ref{fig:figure1}.
A user browses different types of pages and then requests a live chat agent.
From the conversation, we derive two types of ground-truth intent: \textit{raw intent} (or simply, \textit{intent}) and \textit{intent class}.
An intent is the collection of a user's raw utterances in the chat and represents the precise reason the user is seeking help, e.g., `Do you provide installation services for this ceiling fan?'
The intent class is a coarse categorization of an intent, e.g., Installation.
Our goal is to generate the raw intent, since it provides more detailed information such as the item in question and the specific issue being faced.

\begin{figure*}[t]
    \centering
    \vspace{1em}
    \includegraphics[width=0.95\linewidth]{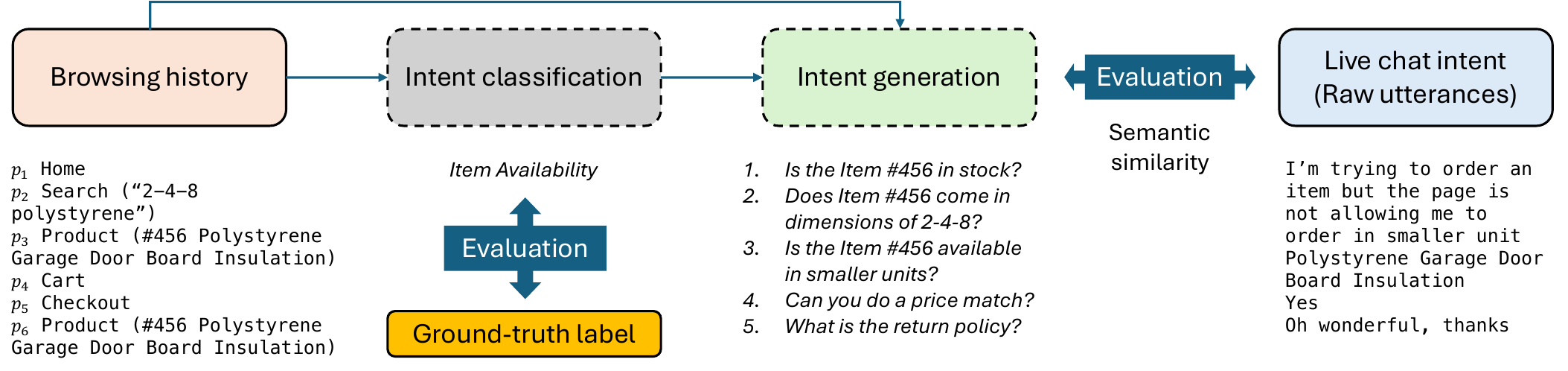}
    \caption{Method overview. The components with dashed borders represent our two-stage approach. Intent classification is trained and evaluated using ground-truth class labels; intent generation is evaluated based on raw intents.}
    \label{fig:method}
\end{figure*}

We propose a two-stage approach as shown in Figure~\ref{fig:method}. 
First, we train a classifier using the intent class labels, where we use pretrained Transformers to capture the rich text semantics in browsing history.
Particularly, we find that the fine-tuned Longformer~\cite{Beltagy2020Longformer} and our variant perform much better than larger text-to-text models (fine-tuned Flan-T5-Large and zero-shot GPT-3.5). 
Second, we use the predicted class to generate intents.
Here, we instruct GPT-3.5 to generate intent candidates from a user's browsing history and the predicted class.
For automatic comparison of ground-truth and generated intents, we use GPT-4 to compare their semantic similarity.
This evaluation method yields a precision of 91\% in alignment with human judgment.
Experiments show that our method achieves significant performance gain compared to generating intents directly without the classification stage.
% Experiment results show that our method achieves a performance gain of twice that of generating intents directly without the classification stage.

%% file: sections/dataset.tex
\section{Dataset}
\label{sec:dataset}

Our dataset was acquired through a collaboration with a major U.S.-based online home improvements retailer, sampled from February 2024.
% The statistics are in Table~\ref{table:statistics}.
There are 30,739 user sessions, with an average of 68 ($\pm 111$) pages per session and 59 ($\pm 75$) words per raw intent.

\boldheading{Live Chat Intents}
When a user clicks on the `chat' button that appears in any page of the website, they are presented with options on their topic of inquiry that may guide them to an automated response.
Still, a user may click on the `chat with an associate' option for personalized assistance, which connects them to a human agent.
We select the chat sessions where the user was connected to a human agent.
Based on the initial option clicks and chat utterances, we assign each chat a single intent class among Installation (INS), Item availability (AVL), Price match (PRI), Repair/Warranty (WTY), and Return/Refund (RET).\footnote{Proportion of sessions: INS (37.3\%), AVL (22.3\%), PRI (21.7\%), WTY (10.4\%), RET (8.2\%)}
These classes cover a majority of intents.
The full intent is often conveyed through multiple exchanges, e.g.,

\begin{description}[labelindent=0pt, font=\normalfont\itshape\space]
    \item[User:] {Hi}
    \item[Agent:] {How may I help you?}
    \item[User:] {Can I get a price match?}
    \item[Agent:] {Could you please provide the item number?}
    \item[User:] {Item \#123456}
\end{description} 
We concatenate the user utterances and use the resulting string as the raw intent (i.e., `Hi, Can I get a price match?, Item \#123456').\footnote{All personal information is substituted with generic placeholders to ensure anonymity.}

% \begin{table}[t!]
% \centering
% \caption{Dataset statistics.}
% \vspace{-1em}
% \label{table:statistics}
% \begin{tabular}{lr}
% \toprule
% \# users (sessions) & 30,739 \\
% \# pages per user & 68 ($\pm 111$) \\
% \# words in raw intent & 59 ($\pm 75$) \\
% Intent classes & INS (37.3\%), AVL (22.3\%), \\
%                & PRI (21.7\%), WTY (10.4\%), RET (8.2\%) \\
% \bottomrule
% \end{tabular}
% \vspace{-1em}
% \end{table}

\boldheading{Browsing Histories}
Each chat session is preceded by the user's browsing history.
We collect the sequence of web pages viewed by the user, each with attributes.
A common attribute across all pages is the page type, which can be one of 66 categories.
The following 9 page types have additional attributes: 
\textit{`product'} (product name); \textit{`search'} and \textit{`products list'} (search query); \textit{`brand'} (brand name); \textit{`catalog'}, \textit{`how to'}, \textit{`buying guide'}, \textit{`inspiration'}, and \textit{`calculators'} (page title).
Sequences with fewer than five pages are discarded, and the rest are split into train (80\%), validation (10\%), and test (10\%) sets.

% \begin{figure}[t]
% \centering
% \begin{mdframed}
% `addressbook', 'appliance delivery preparation', 'best sellers', 'brand', 'buy again', 'buying guide', 'calculators', 'cart', 'catalog', 'checkout', 'collection', 'communications', 'configurable product', 'credit', 'deck designer planner', 'diy projects', 'download-apps', 'gift cards', 'help', 'home', 'how to', 'inspiration', 'installation', 'installation status', 'job applicant privacy', 'license certification', 'login', 'lowest price guarantee', 'loyalty', 'military', 'myaccount', 'mylist', 'myrewards', 'order status', 'orders', 'organization', 'other', 'payment methods', 'privacy statement', 'pro', 'product', 'productComparison', 'products list', 'purchase authorization', 'quotes', 'reco', 'register', 'rental', 'room visualizer', 'saving', 'search', 'shop custom', 'shop events', 'shop flooring samples', 'shop garden center', 'shop online paint', 'shop repair workshops', 'shop tilestudio', 'shopbyroom', 'store', 'store services', 'subscription', 'terms conditions', 'wallet promotions', 'weekly ad', 'workshops'
% \end{mdframed}
% \caption{List of all page types.}
% \label{fig:page-types}
% \vspace{-0.3cm}
% \end{figure}

%% file: sections/methods.tex
\section{Methods}

\subsection{Problem Formulation}
\label{sec:problem}

For each user $u$, we have a browsing history represented by a sequence of pages $S^u = (p^u_1, p^u_2, \dots, p^u_{|S^u|})$.
Each page $p^u_t$ browsed by user $u$ at time $t$ is represented by an attribute dictionary $D_{p^u_t}$ of key-value pairs $\{(k_1, v_1), (k_2, v_2), \dots, (k_l, v_l)\}$, where $k$ denotes the attribute name (e.g., \textit{page type}) and $v$ the corresponding value (e.g., \textit{product page}).
A user is also associated with a natural-language intent $I_u$ and its intent class $c_u = c(I_u) \in C$, where $C$ is the set of intent classes.
Given the sequence $S^u$, the goal is to generate intent $\hat{I}_u$ that is semantically similar to the true intent $I_u$.
We use a similarity function $\chi: (I_u, \hat{I}_u) \rightarrow \{0,1\}$, which returns 1 if the intent pairs are similar and 0 otherwise.
% Given the sequence $S^u$, the goal is to generate the intent $I_u$.
% In practice, it is infeasible to generate the exact $I_u$, since countless different forms may express the same meaning.
% Specifically, our goal is to generate intent $\hat{I}_u$ that is semantically similar to $I_u$, based on the similarity function $\chi: (I_u, \hat{I}_u) \rightarrow \{0,1\}$, which returns 1 if the intent pairs are similar and 0 otherwise.

\subsection{Two-Stage Intent Prediction}

Figure~\ref{fig:method} shows our proposed method, which consists of two intent prediction stages: classification and generation.

\begin{figure}[t]
    \centering
    \vspace{1em}
    \includegraphics[width=0.9\linewidth]{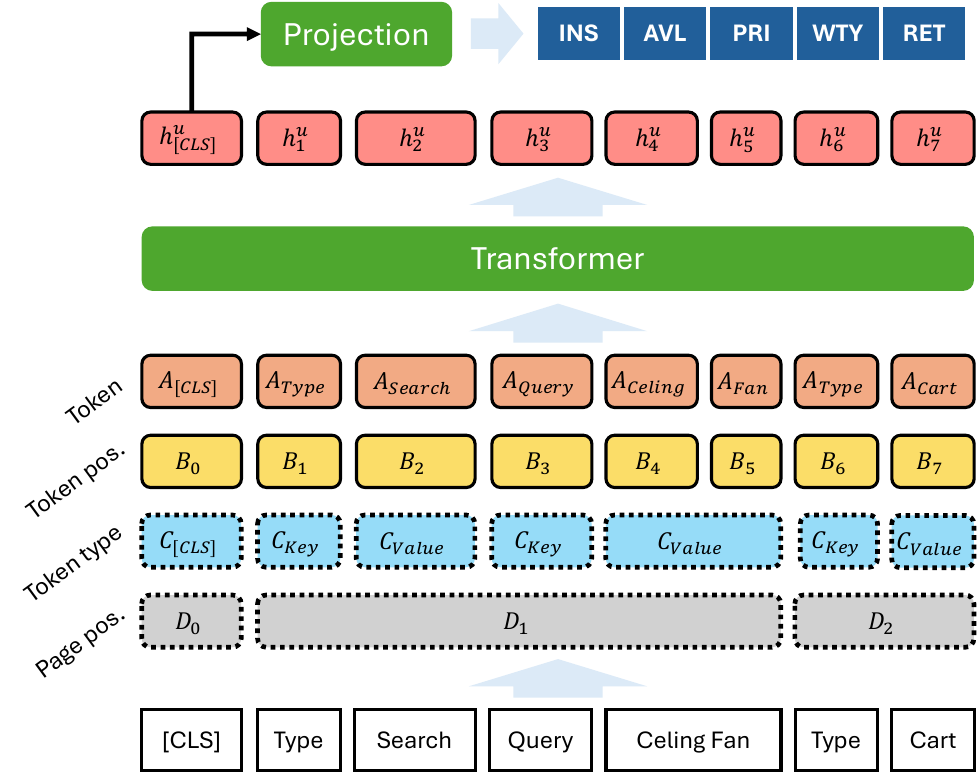}
    \caption{Longformer+ model architecture for intent classification. Removing the optional embeddings (dotted lines) becomes the original Longformer.}
    \label{fig:xformer}
\end{figure}

\boldheading{Intent Classification}
The first stage learns a classifier $f$ to predict the intent class $c_u$ from the browsing history: $f: S^u \rightarrow \hat{c}_u$.
Since each browsing history is represented by a sequence of dictionaries, we build a model tailored to this structure, inspired by~\citet{li2023text}.
Figure~\ref{fig:xformer} illustrates this architecture, characterized by four embedding matrices:
token embeddings $\mathbf{A} \in \mathbb{R}^{V_w \times d}$ represent tokens (where $V_w$ is the vocabulary size), token position embeddings $\mathbf{B} \in \mathbb{R}^{p \times d}$ represent the token positions in a sequence (where $p$ is the maximum number of tokens in a sequence), token type embeddings $\mathbf{C} \in \mathbb{R}^{3 \times d}$ indicate one of the three token types ([CLS], key, or value), and page position embeddings $\mathbf{D} \in \mathbb{R}^{n \times d}$ represent the position of pages in a sequence (where $n$ is the maximum number of pages in a sequence).
The embeddings are summed and fed into the transformer layer, where the representation $h_{[CLS]}$ of the [CLS] token is used for prediction through a linear projection layered followed by softmax.
We refer to this model as \textbf{Longformer+}, since it extends the \textbf{Longformer}~\cite{Beltagy2020Longformer} architecture by adding token type and page position embeddings.
The common components are initialized with pretrained parameters.\footnote{https://huggingface.co/allenai/longformer-base-4096}
Maximum number of tokens (1024), pages (50), attribute tokens (32) are fixed and other hyperparameters are obtained through grid search.
Sequences are truncated from the beginning to prioritize recent interactions.
As baselines, we use \textbf{Flan-T5} (large, 780M)~\cite{chung2024scaling} fine-tuned with our training data and \textbf{GPT-3.5} (gpt-3.5-turbo-0125)\footnote{For deterministic results, we set temperature to 0 in all our experiments.}~\cite{openai2023gpt35turbo} without any fine-tuning.
These text-to-text models have larger sizes and more extensive training datasets.
For these models, we flatten the browsing history and mark the beginning of each page with a <page> token.
We use the instruction `Predict the customer's intent behind reaching out to a live chat agent, after viewing a sequence of the following pages:', which is prepended to the browsing history.
Each output text is matched to one of the five classes.

\boldheading{Intent Generation}
The second stage instructs a natural language generator $g$ to output intent $I_u$ from browsing history and the predicted intent class: $g: (S^u, \hat{c}_u) \rightarrow \hat{I}_u$.
Precisely, we instruct $g$ to generate $M$ intent candidates: $(\hat{I}_u^1, \hat{I}_u^2, ..., \hat{I}_u^M)$. 
We use \textbf{GPT-3.5}\footnote{We use zero-shot GPT-3.5, which is significantly more cost-effective than fine-tuning it. Although we explored fine-tuning smaller models such as Flan-T5, the outputs did not make much sense.} as our generator to leverage its commonsense reasoning capabilities~\cite{brown2020language, li2021systematic}. 
For instance, if a user visited a product page and a checkout page, the user may be considering  purchasing this item.
The predicted intent class $\hat{c}_u$ obtained from the trained classifier $f$ provides further guidance based on learned patterns that may not be evident to zero-shot models.
We use the following instruction:
`A customer browsed the following pages---$S^u$---and reached out a chat agent for assistance. Possible topics are $C$, but $\hat{c}_u$ is the most likely. Pretend to be this customer, and enumerate $M$ questions (1., 2., ...) to ask the chat agent. Don't say anything else.' 
Here, the browsing history $S_u$ is flattened with a <page> token (as in  classification), and $C$ is randomly shuffled to mitigate positional bias~\cite{zheng2023large}.
As baselines, \textbf{Use All} omits $\hat{c}_u$ but still states that the topic is one of $C$; \textbf{Use None} excludes both $\hat{c}_u$ and $C$.

\subsection{Evaluation}
We evaluate whether among the $M$ generated intent candidates, there is one that is semantically similar to the true intent $I_u$.
Doing so requires a similarity function $\chi$ (see Section~\ref{sec:problem}).
While a human expert is the ideal judge, we want an inexpensive, automated alternative. 
Traditional count-based~\cite{lin2004rouge, papineni2002bleu} or embedding-based~\cite{reimers2019sentence} similarity metrics would be unsuitable: (1) these metrics output scores on a scale, rather than binary values, (2) they do not handle `dealbreakers' such as different item names, and (3) they may overlook the subtle but important difference in intents (e.g., asking for installation manual versus asking for installation services).
Hence, we propose a new evaluation scheme: use a large language model (LLM) as a similarity function.
Particularly, we prompt GPT-4 (gpt-4-0125-preview)~\cite{openai2023gpt4turbo} with instructions (to determine if the two intents are similar), including demonstrations.
We validate GPT-4 judgments by having human workers assess 200 intent pairs, achieving Cohen's kappa of 0.71, precision of 91.8\%, and recall of 82.6\%.
These results suggest that the GPT-4 judgments are reliable and slightly less generous than humans in determining similarity.

%% file: sections/results.tex
% \begin{table}[t!]
% \centering
% \caption{\todo{Caption}}
% \label{table:classification}
% \begin{tabularx}{\columnwidth}{llcccccc}
% \toprule
% & & & \multicolumn{5}{c}{\textbf{Intent class}} \\
% \cmidrule(lr){4-8}
% \textbf{Metric} & \textbf{Model} & \textbf{All} & \textbf{INS} & \textbf{AVL} & \textbf{PRI} & \textbf{WTY} & \textbf{RET} \\
% \midrule
% Precision & Long+  & \bf .683 & \bf.750 & \underline{.701} & \underline{.784} & \bf.726 & .514 \\
%         & Long     & \underline{.646} & \underline{.685} & \bf.714 & .769 & \underline{.429} & \underline{.540} \\ 
%         & Flan-T5  & .556 & .666 & .555 & \bf.790 & .0 & \bf.623 \\ 
%         & GPT-3.5  & .447 & .538 & .424 & .495 & .254 & .462 \\ 
% \midrule
% Recall  & Long+      & \bf .658 & \underline{.685} & .743 & \underline{.555} & \underline{.116} & \bf.778 \\
%         & Long      & \underline{.649} & \bf.714 & .564 & \bf.564 & \bf.262 & \underline{.653} \\
%         & Flan-T5   & .572 & .509 & \bf .858 & .449 & .0 & .455 \\ 
%         & GPT-3.5   & .450 & .425 & \underline{.749} & .357 & .094 & .148 \\ 
% \bottomrule
% \end{tabularx}
% \end{table}

\begin{table}[t!]
\centering
\caption{Results for intent classification.}
% \vspace{-0.8em}
\label{table:classification}
\begin{tabularx}{\columnwidth}{llcccccc}
\toprule
& & & \multicolumn{5}{c}{\textbf{Intent class}} \\
\cmidrule(lr){4-8}
\textbf{Metric} & \textbf{Model} & \textbf{All} & \textbf{INS} & \textbf{AVL} & \textbf{PRI} & \textbf{WTY} & \textbf{RET} \\
\midrule
Precision & Long+  & \bf .683 & .750 & {.701} & {.784} & .726 & .514 \\
        & Long     & \underline{.646} & {.685} & .714 & .769 & {.429} & {.540} \\ 
        & Flan-T5  & .556 & .666 & .555 & .790 & .0 & .623 \\ 
        & GPT-3.5  & .447 & .538 & .424 & .495 & .254 & .462 \\ 
\midrule
Recall  & Long+      & \bf .658 & {.685} & .743 & {.555} & {.116} & .778 \\
        & Long      & \underline{.649} & .714 & .564 & .564 & .262 & {.653} \\
        & Flan-T5   & .572 & .509 &  .858 & .449 & .0 & .455 \\ 
        & GPT-3.5   & .450 & .425 & {.749} & .357 & .094 & .148 \\ 
\bottomrule
\end{tabularx}
\end{table}

\begin{table}[t!]
\centering
\caption{Results for fine-grained intent generation. We also show the performance of using a perfect classifier (i.e., using ground-truth class labels) as reference.}
% \vspace{-0.8em}
\label{table:generation}
\begin{tabular}{lcc}
\toprule
& \bf Similar@1 & \bf Similar@5 \\
\midrule
Use Ground-Truth Class & .3333 & .5567 \\
\midrule
Use Predicted Class & \bf .2567 & \bf .4500 \\
Use None & .1333 & .3433 \\
Use All & .1333 & .2900 \\
\bottomrule
\end{tabular}
% \vspace{-0.5em}
\end{table}

{
\renewcommand{\arraystretch}{0.7}
\begin{table*}[t]

\centering
\begin{tabularx}{\textwidth}{S|T}
\toprule
\bf Human intent & \bf Generated intents (GPT-3.5) and evaluation (\autoeval{GPT-4}, \humaneval{human})\\ 
\midrule
\small
    \myvspace
    Hi, I'm interested in potentially purchasing two large appliances. \intent{How do I add installation on these, or is it included?}, 
    \itemtitle{LG ColdWash 5-cu ft Agitator Top-Load Washer (Grey)}
    \itemtitle{LG 7-cu ft Electric Dryer (Grey)},
    wonderful. is it best to call my local store, or place the order online?
 & 
 \small 
    \begin{enumerate}[leftmargin=*, label=\arabic*.]
        \item Can you help me with the \intent{installation process} for the \itemtitle{LG 7-cu ft Electric Dryer}? (\autoeval{\success}, \humaneval{\success})
        \item Do you offer \intent{installation services} for the \itemtitle{Amana 6-cu ft Agitator Top-Load Washer}? (\autoeval{\success}, \humaneval{\failure})
        \item What is the \intent{difference between} the \itemtitle{LG ColdWash 5-cu ft Agitator Top-Load Washer} and the \itemtitle{LG 7-cu ft Electric Dryer}? (\autoeval{\failure}, \humaneval{\failure})
        \item Can you provide \intent{more information} on the \itemtitle{LG 7-cu ft Side Swing Door Gas Dryer}? (\autoeval{\failure}, \humaneval{\failure})
        \item Do you have any \intent{recommendations for accessories or additional products} to go with the \itemtitle{LG 7-cu ft Electric Dryer}? (\autoeval{\failure}, \humaneval{\failure})
    \end{enumerate}
 \\
\midrule
\small 
    \myvspace
    So I purchased a thermostat a few months ago. I loved the item and ordered another online. It was the wrong one. Today I ordered another one like the first one. It is \itemtitle{TH1234 smart thermostat}.  I can not find the receipt. \intent{Is it possible to return it?}, I want to return \itemtitle{TH5678}.
& 
\small 
    \begin{enumerate}[leftmargin=*, label=\arabic*.]
        \item Can I \intent{return} the \itemtitle{TH1234 thermostat} for a refund? (\autoeval{\failure}, \humaneval{\failure}).
        \item What is the \intent{process for returning} an item for a refund? (\autoeval{\failure}, \humaneval{\success})
        \item Are there any \intent{restocking fees for returning} the \itemtitle{thermostat}? (\autoeval{\success}, \humaneval{\success})
        \item \intent{How long do I have to return} the \itemtitle{thermostat} for a refund? (\autoeval{\failure}, \humaneval{\failure})
        \item Can I \intent{exchange} the \itemtitle{thermostat} for a different model instead of a refund? (\autoeval{\failure}, \humaneval{\failure})
    \end{enumerate}
\\
\midrule
\small 
    \myvspace
    I am looking at this \itemtitle{electric power lift heat and massage recliner}, It says in the \intent{description that it also rocks} but i cannot see how that is possible, I do not want that feature \intent{can you verify if this is correct} also \intent{if I can see it and try it somewhere}, The item is \itemtitle{REC1234}
& 
\small 
    \begin{enumerate}[leftmargin=*, label=\arabic*.]
        \item Are there any \itemtitle{lift recliners} currently \intent{in stock}? (\autoeval{\failure}, \humaneval{\failure})
        \item Can you provide me with \intent{more information} on the \itemtitle{lift recliners} listed on the website? (\autoeval{\success}, \humaneval{\success})
        \item Do you have any \intent{recommendations} for \itemtitle{lift recliners} that would suit a small space? (\autoeval{\failure}, \humaneval{\failure})
        \item Are there any \intent{ongoing promotions or discounts} for \itemtitle{lift recliners}? (\autoeval{\failure}, \humaneval{\failure})
        \item Can you help me with the \intent{dimensions} of the \itemtitle{lift recliners} available on your website? (\autoeval{\failure}, \humaneval{\failure})
    \end{enumerate}

\\
\bottomrule
\end{tabularx}
\myvspace
\caption{Examples of intents generated by our two-stage approach. We highlight \itemtitle{items} and \intent{keyphrases} for better comparison. GPT-4 and human evaluation (\success\space if intents are evaluated as similar and \failure\space otherwise) align overall.}
\label{tab:cases}
\vspace{-2em}
\end{table*}
}

\section{Results}

\subsection{Classification Results}

Table~\ref{table:classification} shows the classification results on the test set. 
We measure precision and recall for each class and compute the weighted averages based on class proportion as overall performance.
Longformer+ achieves the best overall results, with its precision showing gains of 5.7\%, 22.8\%, and 52.8\% over Longformer, Flan-T5, and GPT-3.5, respectively.
Intuitively, if the output of Longformer+ is not Return/Refund (RET), there is approximately 75\% chance that the class is correct. 
Models may find some classes harder to predict than others:
for example, Longformer+ has high precision and recall for Item availability (AVL), but high precision and low recall on Repair/Warranty (WTY).
The lower overall performance of Flan-T5 and GPT-3.5 could be due to them being general-purpose language models with text as output,
which exhibit inherent output biases~\cite{yoon2024evaluating}, even with fine-tuning in the case of Flan-T5.

\subsection{Generation Results}
\label{sec:generation_results}

Table~\ref{table:generation} shows generation results, where Similar@$m$ indicates the proportion of users where the model generated a similar intent within the top-$m$ list.
Our proposed method achieves rates of 0.2567 ($m=1$) and 0.45 ($m=5$), which are 93\% and 31\% gains over using no class information (Use None).
Interestingly, using no class (Use None) performs better than using all classes (Use All), where the latter informs the model of the five possible classes.
% Examples of generated intents and evaluation results are in Table~\ref{tab:cases}.
We observe that our method successfully generates fine-grained intents similar to the raw intents (examples in Table~\ref{tab:cases}).
Note that the model does not trivially repeat the given intent class (e.g., Item Availability $\rightarrow$ Is the item available?), but generates various inquiries related to the topic.
The model often identifies the correct item, but it sometimes selects the wrong one (e.g., TH1234 instead of TH5678) from the browsing history.
As future work, we plan to improve performance by training a component that selects items.

%% file: sections/related.tex
\section{Related Work}
\label{sec:related}

Intent prediction is an important problem in e-commerce, with downstream applications in user understanding~\cite{wu2015neural}, targeted marketing~\cite{li2020spending}, and recommendation~\cite{li2022automatically}.
Existing work focuses on purchase intent~\cite{guo2019buying, vieira2015predicting, sheil2018predicting, wu2015neural}, where the goal is to classify whether a user would buy something or not.
User sessions include clicks~\cite{sheil2018predicting, wu2015neural},
touch-interactive behaviors~\cite{guo2019buying}, and event types~\cite{vieira2015predicting}.
Some work predicts the specific item a user will purchase after identifying purchase intent~\cite{volkovs2015two}.
Different from these work, our paper generates fine-grained intent in natural language;
our prediction space is not a finite set of classes or items, but free-form text.

A related task is recommendation, where intent is not the direct output but modeled through model architecture~\cite{guo2022learning}, pattern mining~\cite{NEURIPS2023_621d0fd4}, and fine-grained signals~\cite{meng2020incorporating}.
Another line of work infers intent in dialogs~\cite{cai2020predicting, ledneva2024reimagining} without using user behaviors prior to chat.

%% file: sections/conclusion.tex
\section{Further Discussion}

This paper explores a novel problem of using browsing history to forecast why the customer needs a live chat agent.
Our problem introduces new application possibilities, such as recommending actions for users, routing users to specialized agents, and prioritizing intents when agent resources are limited.
We plan to enhance model performance by (1) adding detailed user interactions, (2) creating a component to select items (see Section~\ref{sec:generation_results}), and (3) integrating external knowledge such as FAQs or seasonal sales.